\documentclass[runningheads]{llncs}
\usepackage[T1]{fontenc}
\usepackage{amsmath}
\usepackage{amssymb}
\usepackage{algorithm}
\usepackage{algorithmic}
\usepackage{graphicx}

\usepackage{tikz}
\usetikzlibrary{arrows.meta, positioning}

\begin{document}
	
	\title{Neuro-Evolved Heuristics for Variable Gapped Common Subsequence Identification}
	\titlerunning{Neural-evolved Heuristics for the VGLCSP}
	
	\author{Marko Djukanović\inst{1, 5}  
		\and
	  Christian Blum\inst{2}  \and 
	  Aleksandar Kartelj\inst{3}  \and
	  Sašo Džeroski\inst{4}  \and
	  Žiga Zebec\inst{5}
	}
	\authorrunning{Djukanovic et al.}
	
	\institute{ University of Nova Gorica, Nova Gorica, Slovenia \\ \email{marko.dukanovic@ung.si} \and 
		Artificial Intelligence Research Institute (IIIA-CSIC), Barcelona, Spain \\
		\email{christian.blum@iiia.csic.es} \and 
		Faculty of Mathematics, University of Belgrade, Belgrade, Serbia \\ \email{kartelj@matf.rs} \\ \and
		Jožef Stefan Institute, Ljubljana, Slovenia \ \\ \email{saso.dzeroski@ijs.si}  \\
		\and
		Institute of Information Sciences (IZUM), Maribor, Slovenia \\
		\email{ziga.zebec@izum.si}
	}

	\maketitle              
	
\begin{abstract}

This study addresses the Variable Gapped Longest Common Subsequence Problem (VGLCSP), a variant of the classical longest common subsequence problem with additional gap constraints and applications in sequence alignment and time-series analysis. While the two-sequence version has been widely studied using dynamic programming, the generalized multi-sequence form is usually solved with beam search-based heuristics, whose hand-crafted designs often lack robustness.

To overcome this limitation, we propose a learning-based approach for automatically designing more effective data-driven heuristics. The heuristics are represented by a neural network with predefined architecture, whose weights are optimized by a genetic algorithm within a neuro-evolutionary framework. The learning process alternates between weight optimization and evaluation within an iterative multi-source beam search procedure, a state-of-the-art method for the problem. Rather than constructing solutions directly, the neural network learns to guide the search process, producing a neuro-evolved heuristic. We further introduce an ensemble heuristic that combines the scores of learned and the best-performing hand-crafted heuristic. Integrated into the iterative multi-source beam search framework, the resulting hybrid approach outperforms existing methods on both synthetic benchmark instances and newly introduced real-world instances with data-driven gap constraints.
\end{abstract}


\section{Introduction}

The \textit{Longest Common Subsequence Problem} (LCSP)~\cite{DJUKANOVIC2020106499} is a fundamental combinatorial optimization problem with numerous applications in computational biology, particularly in the structural analysis of molecular sequences. Given a set of input sequences $S=\{s_1,\ldots,s_m\}$ over an alphabet $\Sigma$, the objective is to determine the longest subsequence common to all sequences $s_i \in S$. Over the past decades, several practically motivated variants of LCSP have been proposed by incorporating additional structural or biological constraints, including the constrained, arc-preserving, and repetition-free variants~\cite{lafond2023longest,lin2002longest}.

In this work, we study the \textit{Variable Gapped Longest Common Subsequence Problem} (VGLCSP), introduced in~\cite{penga2011longest}, theoretically examined in~\cite{adamson2023longest}, and recently experimentally evaluated  in~\cite{djukanovic2026_eurocast}. This variant extends the classical LCSP by introducing flexible gap constraints between consecutive symbols in the subsequence, allowing these gaps to vary along the symbols of input sequences.

Formally, for each sequence $s_i \in S$, a gap function $G_{s_i} : \{1,\ldots,|s_i|\} \rightarrow \mathbb{N}$ specifies the maximum allowed distance between consecutive symbols of the resulting sequence and their positions in the input sequences. If a solution $s$ appears at positions $j_i^1 < \cdots < j_i^{|s|}$ in $s_i$ as a subsequence, the gap constraints are satisfied if $
j_i^x - j_i^{x-1} \leq G_{s_i}(j_i^x) + 1, \quad \forall x = 2,\ldots,|s|.$ 
A subsequence $s$ is feasible if these constraints hold for all sequences $s_i \in S$. The objective is to find the longest feasible subsequence. The VGLCSP provides a flexible and realistic model for sequence alignment, particularly relevant for DNA and protein analysis with variable structural distances, as well as for time-series analysis where events must occur within specified temporal delays~\cite{lainscsek2015delay}.

While several exact dynamic programming approaches exist for the two-sequence case ($m=2$)~\cite{peng2014finding}, the problem becomes computationally intractable for larger $m$; in fact, it is NP-hard for arbitrary $m$; no efficient exact method is known. The only available reasonably scalable approach is an iterative multi-source beam search (IMSBS)~\cite{djukanovic2026_eurocast}, which explores a state-space representation of the problem. The method iteratively refines candidate root nodes and extends them through feasible symbol additions. Despite its scalability, IMSBS relies on hand-crafted heuristics, which tend to degrade in performance as $m$ increases or the alphabet size $|\Sigma|$ decreases~\cite{DJUKANOVIC2020106499}. The presence of gap constraints further exacerbates this issue.  In addition, existing studies are limited to synthetic benchmark instances only, leaving a gap in evaluating methods on realistic data.

To address these limitations, we propose a learning-enhanced beam search framework for VGLCSP. Our approach integrates learned and hand-crafted heuristics into a unified search strategy into a better search guidance. The learned heuristic is modeled by a multi-layer neural network that scores candidate states---that is, feasible extensions of partial solutions---during the search. The network parameters are optimized through a genetic algorithm in a neuro-evolutionary framework, where each candidate model is evaluated within the IMSBS. Importantly, the neural model does not directly construct solutions but learns to guide the search process. Furthermore, we design an ensemble heuristic that combines the learned heuristic with the best-performing hand-crafted heuristic, yielding a more robust guidance mechanism. This hybrid approach is inspired by recent advances in learning-guided combinatorial optimization~\cite{djukanovic2025learning,reixach2024neural,wu2021learning}, but differs in explicitly leveraging complementary heuristic scores within an ensemble.

Finally, to address the lack of realistic benchmarks, we introduce new problem instances derived from biologically motivated datasets from LCSP, augmented with biologically-motivated data-driven gap constraints.\\

\noindent\textit{Contributions.} The main contributions of this work are:
\begin{itemize}
	\item A neuro-evolutionary framework designed for learning heuristics (resp.~heuristic guidance) in the context of the VGLCSP;
	\item An ensemble heuristic combining learned and hand-crafted guidance;
	\item An enhanced IMSBS algorithm integrating the proposed ensemble heuristic;
	\item A new set of realistic benchmark instances with data-driven gap constraints.
\end{itemize}

\paragraph{Preliminaries.}	Let $|s|$ denotes the length of a sequence $s$, and let $s[i]$ refers to the character of $s$ at position $i$, where indexing starts at $i = 1$. The substring of $s$ that begins at position $i$ and ends at position $j$ is denoted by $s[i,j]$. If $i = j$, this corresponds to the single character $s[i]$; if $j < i$, then $s[i,j] = \varepsilon$, where $\varepsilon$ denotes the empty string.	Given two sequences $s_1$ and $s_2$, by $s_1 \cdot s_2$ we denote the sequence obtained by concatenating the given two sequences. 
The number of occurrences of letter $a \in \Sigma$ in $s$ is denoted by $|s|_a$.	We denote by $S = \{s_1, \ldots, s_m\}$ the set of input sequences, $m \in \mathbb{N}$ the number of input sequences (and, correspondingly, the number of gap constraints), and $n$ denotes the length of the longest sequence in $S$.  Given a positional vector $\mathbf{p}^L \in \mathbb{N}^m$, where $1 \leq p^L_i \leq |s_i|$ for all $s_i \in S$, the subproblem related to these positions is given by $S[\mathbf{p}^L]=\{ s_i[\mathbf{p}^L_i, |s_i|] \mid i=1, \ldots, m\}$. \\
  
The paper is organized as follows. In Section~\ref{sec:imsbs}, the iterative multi-source beam search approach from the literature is roughly presented. Section~\ref{sec:ffnn_guidance} provides the main details on the process of learning heuristic guidance and the design of an ensemble heuristic estimator for the tackled problem. Section~\ref{sec:experiments} reports a rigorous experimental analysis. Finally, Section~\ref{sec:conclusions} concludes the paper, highlighting several future research directions.

\section{Iterative Multi-Source Beam Search for the VGLCSP}\label{sec:imsbs}

The first scalable approach for the VGLCSP with an arbitrary number of sequences ($m$) was introduced in~\cite{djukanovic2026_eurocast} as an \textit{Iterative Multi-Source Beam Search} (IMSBS). In this section, we summarize its main components; interested readers are referred to the original paper to access the  full details.

IMSBS is based on a \textit{state graph} representation of the VGLCSP. Each node represents a partial feasible subsequence, characterized by a vector $\mathbf{p}^{L,v}$ of positions defining a subproblem and the length $l^v$ of the constructed subsequence. Formally, a state $v=(\mathbf{p}^{L,v}, l^v)$ is induced by a partial solution $s^v$ if: (i) $p^{L,v}_i-1$ is the smallest index such that $s^v$ is a subsequence of the prefix $s_i[1, p^{L,v}_i-1]$; (ii) $l^v=|s^v|$; and (iii) all gap constraints $G_{s_i}$ are satisfied w.r.t.\  $s^v$. 

A directed arc $\alpha=(v_1, v_2)$ labeled by $lett(\alpha)=a \in \Sigma$ exists if $l^{v_2}=l^{v_1}+1$ and $s^{v_2}=s^{v_1}\cdot a$. To expand a state $v$, all letters occurring in the suffixes $S[\mathbf{p}^{L,v}]$ are considered. For each such letter $a$, the smallest feasible positions $p^{L,a}_i \geq p^{L,v}_i$ are identified such that the gap constraints are satisfied, i.e., $p^{L,a}_i - p^{L,v}_i \leq G_{s_i}(p^{L,a}_i)$. The resulting child state (if not dominated) is given by $w=(\mathbf{p}^{L,a}+\mathbf{1}, l^v+1)$. Terminal states correspond to non-extendable feasible subsequences.

\begin{figure}[!ht]
	\centering
	\scalebox{0.6}{
		\begin{tikzpicture}[
			node distance=1.8cm,
			every node/.style={
				draw=blue!60!black,
				fill=blue!5,
				rectangle,
				rounded corners,
				minimum width=2.8cm,
				minimum height=0.8cm,
				align=center
			},
			arrow/.style={->, thick, color=blue!70!black},
			optarrow/.style={->, ultra thick, color=blue!90!black}, 
			edgelabel/.style={midway, right, draw=none, fill=none, text=black}
			]
			
			\node (r1) at (0,0) {$((1,1),0)$};
			\node (a1) [below of=r1] {$((2,2),1)$\\\small A};
			\node[fill=lightgray] (t1) [below of=a1] {$((3,3),2)$\\\small AT};
			
			\draw[arrow] (r1) -- node[edgelabel] {A} (a1);
			\draw[arrow] (a1) -- node[edgelabel] {T} (t1);
			
			\node (r2) at (6,0) {$((5,5),0)$};
			\node (a2) [below of=r2] {$((6,6),1)$\\\small A};
			\node (a3) [below of=a2] {$((7,7),2)$\\\small AA};
			\node (a4) [below of=a3] {$((8,8),3)$\\\small AAA};
			\node[fill=lightgray] (a5) [below of=a4] {$((9,9),4)$\\\small AAAA};
			
			\draw[optarrow] (r2) -- node[edgelabel] {A} (a2);
			\draw[optarrow] (a2) -- node[edgelabel] {A} (a3);
			\draw[optarrow] (a3) -- node[edgelabel] {A} (a4);
			\draw[optarrow] (a4) -- node[edgelabel] {A} (a5);
			
			\node[draw=none, right=0.2cm of a5] {\small optimal};
			
			\node[draw=none, above=0.8cm of r1] {\textbf{Component 1 (from $(1,1)$)}};
			\node[draw=none, above=0.8cm of r2] {\textbf{Component 2 (from $(5,5)$)}};
			
	\end{tikzpicture}}
	\caption{State graph for $s_1=\texttt{ATGGAAAA}$ and $s_2=\texttt{ATCCAAAA}$ with $G_{s_1}=G_{s_2}=1$. The state graph consists of two disconnected components. Starting from $(1,1)$ leads to a suboptimal solution, while starting from $(5,5)$ yields the optimal subsequence $\texttt{AAAA}$.}
	\label{fig:state_graph_vglcs_example}
\end{figure}
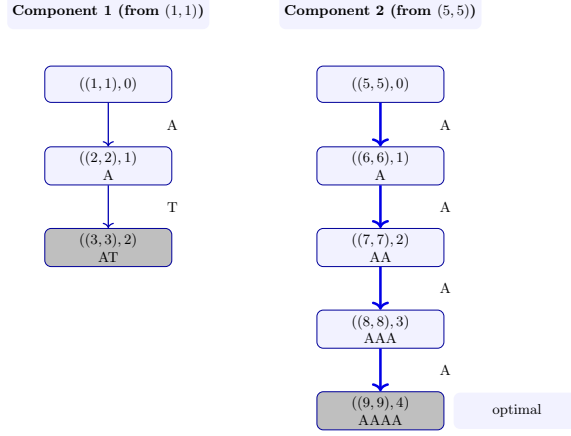

Unlike the classical LCSP, the VGLCSP admits multiple (potentially exponentially many) root nodes, since starting positions are unconstrained. This may lead to disconnected regions in the state graph. For instance, consider $s_1=\texttt{ATGGAAAA}$ and $s_2=\texttt{ATCCAAAA}$ with $G_{s_1}=G_{s_2}=1$. The root node $((1,1),0)$ cannot reach states with position vector $(5,5)$, thus missing the optimal subsequence $\texttt{AAAA}$, generated from $(5,5)$. This phenomenon is illustrated in Fig.~\ref{fig:state_graph_vglcs_example}, where the state graph decomposes into (two) disconnected components.

This observation motivates the IMSBS framework, which operates over a dynamically generated set of root nodes explored iteratively. At each iteration, beam search (BS) is applied to a set of selected root nodes, seeking for complete solutions. BS is a breadth-first heuristic search strategy~\cite{wiseman2016sequence} that expands a limited number of nodes per level, controlled by the beam width $\beta$, and guided by a heuristic function $h$.

In the original IMSBS, node selection relies on an upper-bound heuristic defined as $
\texttt{UB}(v)= l^v + \sum_{a \in \Sigma} \min_{i=1,\ldots,m} |s_i[p^{L,v}_i, |s_i|]|_a,$ 
which estimates the maximum achievable extension from state $v$. The heuristic evaluates remaining occurrences of each letter $a\in\Sigma$ in each suffix sequence of the related subproblem and aggregates these values to derive an upper bound on the achievable subsequence length.

Despite its scalability, IMSBS relies on hand-crafted heuristics whose effectiveness degrades in challenging settings (e.g., large $m$ or small alphabets), especially under gap constraints. {To resolve the issue, we will develop a learning-based methodology that enhances the search process of the IMSBS through the integrated learned heuristic guidance. }
 
	In the subsequent part of the paper, we refer to the algorithm as \texttt{IMSBS}($\mathcal{I}$, $\mathcal{C}$),  
	returning an approximate solution $s_{imsbs}$ for an instance $\mathcal{I}$ under the parameter configuration $\mathcal{C}$.  

\section{Learning to Guide the Search} \label{sec:ffnn_guidance}

Neural networks (NNs) are well-established machine learning models that have demonstrated strong performance across a wide range of domains~\cite{gurney2018introduction,bishop1994neural}. They range from simple multi-layer perceptrons (MLPs) to more complex convolutional and recurrent architectures~\cite{mienye2024recurrent}. In this work, we focus on fully connected multi-layer perceptrons to design an alternative, data-driven heuristic for guiding the IMSBS procedure~\cite{djukanovic2026_eurocast}. 

The key challenge lies in parameterizing the neural model such that it provides meaningful guidance during the search. Specifically, the network is trained to assign a score to each node of the generated state graph---that is, to each feasible extension of a current partial solution---based on features extracted from the corresponding partial solution and the associated subproblem.

In standard neural network training, parameters are optimized via forward and backward propagation, typically using gradient-based methods such as  gradient descent~\cite{amari1993backpropagation}. However, in our setting---characterized by a combinatorial \textsc{NP}-hard problem---target values are not readily available, especially for large instances. This limits the applicability of supervised learning approaches. Instead, we adopt a population-based metaheuristic approach, where the parameters of NN (as configurations) are evaluated indirectly through their impact on the performance of the IMSBS algorithm. 

\medskip
\noindent\textbf{Predefined NN Architecture.}
The heuristic is modeled as a multi-layer perceptron consisting of an input layer, several hidden layers with user-defined sizes, and a single output neuron. The input layer receives a vector of numerical features describing the node (partial solution) to be evaluated, while the output represents the estimated promise of producing that state for further consideration in the IMSBS.

For each layer $l$, forward propagation is defined as
\[
\mathbf{z}^{(l+1)} = W^{(l)} \mathbf{a}^{(l)} + b^{(l)}, \quad
\mathbf{a}^{(l+1)} = \phi(\mathbf{z}^{(l+1)}),
\]
where $W^{(l)}$ and $b^{(l)}$ denote the weights and biases associated with layer $l$, and $\phi$ is a non-linear activation function (e.g., \texttt{tanh}, \texttt{ReLU}, or \texttt{sigmoid}). The final value is used as the heuristic score. All parameters are flattened into a single vector $w$ 
  which represents an individual in the evolutionary process.

\medskip
\noindent\textbf{Fitness Evaluation.}
Given a candidate weight vector $w$, the evaluation proceeds as follows: 
(i) the weights are loaded into the pre-defined NN architecture $\mathcal{N}$; 
(ii) IMSBS is executed on each instance from a training set $\mathcal{T}$, making use of the induced neural heuristic $h_w$ to guide the search (other parameters of IMSBS are fixed, see Section~\ref{sec:experiments}); and 
(iii) the best solution quality obtained per instance is recorded. 
The fitness of $w$ is defined as the average solution quality over $\mathcal{T}$:
\begin{equation} \label{eq:training_problem_optimization}
	\max_{w} \; \emph{Fitness}(w)=\frac{1}{|\mathcal T|} \sum_{I \in \mathcal{T}} f(\texttt{IMSBS}(I, h_w)),
\end{equation}
where $f(\cdot)$ denotes the length of the best subsequence found for running IMSBS on instance $I$ with heuristic $
h_w$. Validation performance is computed analogously on a separate dataset $\mathcal{V}$.

\medskip
\noindent\textbf{Feature Extraction.}
To guide the search effectively, each node $v=(\mathbf{p}^v, l^v)$ is mapped to a fixed-length feature vector capturing both the construction progress and the structure of the related subproblem. The design is instance-independent and robust across varying problem sizes.

Let $S=\{s_1,\ldots,s_m\}$ denote the input sequences. The position vector $\mathbf{p}^v$ is first normalized by sequence lengths: $
F^v_i = \frac{p^v_i}{|s_i|}, \quad i=1,\ldots,m,$ 
ensuring scale invariance. From this normalized vector, we extract four statistics: $\max(\mathbf{F}^v)$, $\min(\mathbf{F}^v)$, $\texttt{mean}(\mathbf{F}^v)$, and $\texttt{std}(\mathbf{F}^v)$, capturing the progress and imbalance across sequences. The current subsequence length $l^v$ is also included.

To reflect future feasibility, we compute gap-related statistics: $
G^{\min}, \; G^{\max},$ $G^{\texttt{avg}}, \; G^{\texttt{std}}, $
evaluated over the gap values for the remaining positions in all sequences. These features quantify the flexibility of future extensions: larger gaps indicate more freedom, while smaller or highly variable gaps suggest tighter constraints.

Finally, two global features—alphabet size $|\Sigma|$ and the number of sequences $m$—are included. All features are standardized to zero mean and unit variance prior to training. The resulting 11-dimensional representation provides a compact yet expressive description of each search state.

\medskip
\noindent\textbf{Optimization of NN Parameters.}
To optimize the NN parameters, we employ a \textit{Biased Random-Key Genetic Algorithm} (BRKGA), a population-based evolutionary method. Each individual encodes a candidate weight vector $w$ of the NN model, initialized uniformly at random in the interval $[-q,q]$ with $q=1$. The evolutionary process iteratively refines the population toward high-quality heuristics $h_w=\mathcal{N}(w)$, where $\mathcal{N}$ is a predefined NN architecture, based on the fitness defined in Eq.~(\ref{eq:training_problem_optimization}). The overall learning mechanism is summarized in Algorithm~\ref{alg:learning_mechanism}.

\begin{algorithm}[!ht]
	\caption{GA-Based Learning of a Beam Search Heuristic} \label{alg:learning_mechanism}
	\begin{algorithmic}[1]
		\STATE \textbf{Input}: NN architecture $\mathcal{N}$, IMSBS parameters, BRKGA parameters, training set $\mathcal{T}$, validation set $\mathcal{V}$
		\STATE \textbf{Output}: optimized weight vector $w^\ast$
		\STATE Initialize population $P$ with random weight vectors \# random NN-based heuristics
		\STATE Evaluate all individuals on training set $\mathcal{T}$
		\STATE $w^\ast \leftarrow$ best individual in $P$
		\WHILE{time limit $t_{\max}$ not exceeded}
		\STATE Sort $P$ in descending order of \emph{Fitness}
		\STATE Initialize new population $P' \gets \emptyset$
		\STATE Copy top $n_{\text{elite}}$ individuals from $P$ to $P'$
		
		\FOR{$i = 1$ to $n_{\text{mutants}}$}
		\STATE Generate random individual $x$
		\STATE Evaluate fitness of $x$ on $\mathcal{T}$ and add to $P'$
		\STATE Update $w^\ast$ if improved and validation does not degrade
		\ENDFOR
		
		\FOR{$i = 1$ to $n_{\text{offspring}}= n_{\text{pop}}- n_{\text{elite}} - n_{\text{mutants}}$}
		\STATE  Select one elite parent $x$ and one non-elite parent $y$ from $P[:n_{\textrm{pop}} - n_{\textrm{mutants}}]$
		\STATE Generate offspring between $x$ and $y$ via biased crossover ($p_{\text{bias}}$)
		\STATE Evaluate offspring on $\mathcal{T}$ and add to $P'$
		\STATE Update $w^\ast$ if improved and validation does not degrade
		\ENDFOR
		\IF{all candidates in $P'$ with better fitness than $w^*$ have a degraded validation score }
		    \STATE \textbf{break} \# over-fitting detected
		\ENDIF
		\STATE $P \leftarrow P'$
		\ENDWHILE
		\STATE \textbf{return} $w^\ast$
	\end{algorithmic}
\end{algorithm}

 \textbf{Overall Learning Procedure.}
 The proposed learning framework alternates between evolutionary optimization of neural network parameters and their evaluation within the IMSBS. Importantly, the neural network does not directly construct solutions; instead, it learns to \emph{guide} the search process, placing the approach within the class of neuro-evolutionary hyper-heuristics~\cite{elsaid2019ant,ortiz2016neuro,galvan2021neuroevolution}.
 
 Given a predefined NN architecture $\mathcal{N}$, IMSBS parameters, and BRKGA parameters, the procedure begins by generating an initial population $P$ of candidate weight vectors. Each individual is evaluated on the training set $\mathcal{T}$ using IMSBS guided by the induced neural heuristic. The best-performing solution is stored as $w^\ast$. In our implementation, IMSBS is executed with parameters $\beta=500$ and $\texttt{beam\_iters}=100$, and the time limit is set to 60 seconds, while the remaining parameters follow~\cite{djukanovic2026_eurocast}. This configuration provides a suitable balance between computational cost and robustness during training.
 
 The evolutionary process proceeds iteratively as follows. The population is first sorted according to the fitness values. A new population $P'$ is then constructed by preserving the top $n_{\text{elite}}$ individuals (elitism), introducing $n_{\text{mutants}}$ randomly generated individuals (diversification), and generating $n_{\text{offspring}}$ offspring via biased crossover. In each crossover, one parent is selected from the elite set and the other from the remaining population. Each gene (node weight) is inherited from the elite parent with probability $p_{\text{bias}} \geq 0.5$, and from the second parent otherwise. Each newly generated individual is evaluated on $\mathcal{T}$. Whenever an improved solution is found, its performance is additionally assessed on the validation set $\mathcal{V}$. To mitigate overfitting, we adopt an early stopping strategy: if the validation performance deteriorates for an individual with better fitness value than the current incumbent, it is prevented from becoming a new incumbent. Additionally, if the validation scores of each such individual deteriorate at an iteration of BRKGA, the search is terminated earlier due to the detected overfitting; otherwise the process repeats until a time limit $t_{\max}$ is reached. The algorithm returns the best weight vector $w^\ast$, corresponding to the neural network $\mathcal{N}(w^\ast)$ that induces the heuristic $h_{w^\ast}$.
 
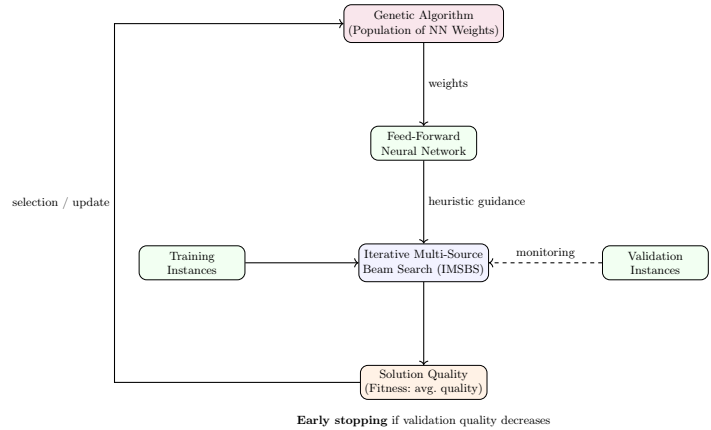
\begin{figure}[!ht]
	\centering
	\scalebox{0.50}{
		\begin{tikzpicture}[
			box/.style={draw, rounded corners=2mm, align=center, minimum width=3.4cm, minimum height=1cm, fill=blue!5},
			smallbox/.style={draw, rounded corners=2mm, align=center, minimum width=2.8cm, minimum height=0.9cm, fill=green!5},
			evalbox/.style={draw, rounded corners=2mm, align=center, minimum width=3cm, minimum height=0.9cm, fill=orange!10},
			arrow/.style={->, thick},
			dashedarrow/.style={->, thick, dashed},
			node distance=2.2cm
			]
			
			\node[box, fill=purple!10] (ga) {Genetic Algorithm\\\small (Population of NN Weights)};
			
			\node[smallbox, below=of ga] (nn) {Feed-Forward\\Neural Network};
			
			\node[box, below=of nn] (bs) {Iterative Multi-Source\\Beam Search (IMSBS)};
			
			\node[smallbox, left=3cm of bs] (train) {Training\\Instances};
			
			\node[smallbox, right=3cm of bs] (val) {Validation\\Instances};
			
			\node[evalbox, below=of bs] (fitness) {Solution Quality\\\small (Fitness: avg.\ quality)};
			
			\draw[arrow] (ga) -- node[right]{weights} (nn);
			\draw[arrow] (nn) -- node[right]{heuristic guidance} (bs);
			
			\draw[arrow] (train) -- (bs);
			\draw[dashedarrow] (val) -- node[above]{monitoring} (bs);
			
			\draw[arrow] (bs) -- (fitness);
			
			\draw[arrow] (fitness.west) -- ++(-6.5,0)          
			|- node[pos=0.25,left]{selection / update} (ga.west);
			
			\node[below=0.3cm of fitness] {\small \textbf{Early stopping} if validation quality decreases};
			
	\end{tikzpicture}}
	
	\caption{Enhanced learning framework integrating a Genetic Algorithm with IMSBS.
		The GA evolves neural network weights, which define a heuristic used within the beam search.
		Training instances guide optimization, while validation instances monitor generalization.
		The feedback loop updates the population based on solution quality.}
	\label{fig:ga_imsbs}
\end{figure}

 Our experiments use the following parameters for BRKGA following~\cite{reixach2024neural}:  $n_{\text{pop}}=20$, $n_{\text{elite}}=1$, $p_{\text{bias}}=0.5$, and $n_{\text{mutants}}=7$. A visualization of the learning process is provided in Fig.~\ref{fig:ga_imsbs}.

  \subsection{An ensemble approach}
    
    During preliminary experiments, we observed that the learned heuristic generally outperforms the hand-crafted heuristic from the literature. However, this advantage is not consistent across all cases, particularly for instances with small values of $m$.  To improve the results further, we employ a rank-based ensemble strategy that combines the learned heuristic with the analytical hand-crafted upper bound~\cite{djukanovic2026_eurocast}, see Section~\ref{sec:imsbs}.  Specifically, each candidate node in beam search is evaluated using two criteria: a neural score obtained from an FNN $\mathcal{N}(w^*)$, denoted $h_{w^*}=\mathcal{N}({w^*})$,  and  a problem-specific hand-crafted $\text{UB}$. 
    
    Instead of combining the raw scores directly, which may differ in scale and distribution, we adopt a rank aggregation approach. First, candidate nodes are sorted independently according to each criterion, producing two rankings. Each node is then assigned a rank based on its position in each sorted list.   The final ranking score is computed as a weighted combination of the two ranks:
    \begin{equation}\label{eq:ensemble_heuristic}
        r(v) = \alpha \cdot r_{h_{w^*}}(v) + (1 - \alpha) \cdot r_{\text{UB}}(v),
    \end{equation}
    Nodes are then sorted in ascending order of $r(v)$, and the top candidates are retained according to the beam width. This rank-based ensemble heuristic provides a robust mechanism for integrating learned and analytical guidance, mitigating scale inconsistencies, and improving search stability. 
     
 
\section{Experimental Evaluation}\label{sec:experiments}

In this section, we evaluate the performance of two approaches: 
(i) \textsc{Imsbs}, the baseline state-of-the-art iterative multi-source beam search from~\cite{djukanovic2026_eurocast}, and 
(ii) \textsc{Limsbs-ensemble}, an enhanced version of \textsc{Imsbs} guided by the ensemble heuristic defined in Eq.~(\ref{eq:ensemble_heuristic}). All methods are implemented in C++ and compiled using the GCC compiler. Experiments are conducted on the VEGA HPC system (IZUM, Maribor, Slovenia), consisting of 960 compute nodes equipped with AMD EPYC 7H12 CPUs at 2.35\,GHz. All experiments are executed in a single-threaded setting, meaning one node was used in single-threaded mode. Each execution (either for training NNs or running any algorithm) assigns  32GB of memory resources.
 The process of  parameters learning for each neural network ($t_{\max}$) is set to 2 CPU hours.

\subsection{Benchmark Instances}

We consider two types of benchmark sets. {Synthetic instances}, denoted by \textsc{Random}, are generated in~\cite{djukanovic2026_eurocast}. For each combination of $n \in \{50, 100, 200, 500\}$, $m \in \{2, 3, 5, 10\}$, and $|\Sigma| \in \{2, 4\}$, 10 instances are available, resulting in a total of 320 problem instances.

{Biologically motivated instances} are generated to
 complement synthetic data. These instances involve data-driven gap constraints derived from biological sequence properties. The goal is to model context-dependent positional importance using thermodynamic stability information, inspired by sequence alignment scoring and weighted subsequence models.  We employ a dinucleotide-based structural model derived from nearest-neighbor thermodynamic interactions~\cite{santalucia1998unified,watkins2005nearest}. The stability of adjacent nucleotides is captured by the function
\[
f(s[i], s[i+1]) = -\Delta G(s[i], s[i+1]),
\]
where $\Delta G$ denotes the free energy associated with the dinucleotide pair. According to~\cite{santalucia1998unified}, the corresponding weights are estimated by the following values:
\begin{align*}
\texttt{DINUC\_WEIGHT} = \{
\texttt{AA}:1.00, \texttt{TT}:1.00,
\texttt{AT}:0.88, \texttt{TA}:0.58,&\\ 
\texttt{CA}:1.45, \texttt{TG}:1.45,
\texttt{GT}:1.44, \texttt{AC}:1.44,&\\  
\texttt{CT}:1.28, \texttt{AG}:1.28,
\texttt{GA}:1.30, \texttt{TC}:1.30,&\\  
\texttt{CG}:2.17, \texttt{GC}:2.24,
\texttt{GG}:1.84, \texttt{CC}:1.84&\}
\end{align*}

To ensure consistency with the directional nature of gap constraints, we define a \emph{causal} positional contribution by
\[
f^s_k =
\begin{cases}
	f(s[k-1], s[k]), & k > 1, \\
	0, & k = 1.
\end{cases}
\]

This formulation enforces a left-to-right dependency structure, aligning with the feasibility condition of the VGLCSP. The positional influence is then computed as: $
I^s(i) = \sum_{k=\max(1,i-r)}^{i} e^{-\lambda (i-k)} \cdot f^s_k,$
where $r$ is the influence radius and $\lambda$ is a decay factor. This captures the cumulative structural context preceding each position.  To normalize influence values, we apply
\[
\hat{I}^{(j)}_i = \frac{I^{(j)}_i - \min(I^{(j)})}{\max(I^{(j)}) - \min(I^{(j)}) + \epsilon} \in [0,1],
\]
and derive gap constraints via $
\text{Gap}_i[j] = w_{\min} + (w_{\max} - w_{\min}) \cdot \hat{I}^{(j)}_i. $

This mapping assigns larger gaps to structurally stable regions and smaller gaps to less stable ones, reflecting their differing flexibility for matching. For producing instances, we set $w_{\min}=1$, $w_{\max}=10$, $r=10$, and $\lambda=0.5$.


\noindent
Moreover, we used the well-known LCS benchmark sets \textsc{Rat} and \textsc{Virus}~\cite{DJUKANOVIC2020106499}. For each sequence in these datasets, gap constraints are generated using the above procedure. We restrict attention to instances with $|\Sigma|=4$, as larger alphabets (e.g., $|\Sigma|=20$) lead to more-or-less trivial solutions under gap constraints and therefore obtaining less informative solutions.

The resulting dataset, denoted \textsc{Real}, consists of 20 instances (10 from each benchmark set). All sequences have the length of 600, while the number of sequences varies from $m=10$ to $m=200$.

 All benchmark sets, neural network configurations, and source codes  are available at   
\url{https://github.com/markodjukanovic90/VGLCS\_Neural\_Heuristics}. 
 
\subsection{Tuning Process}

The parameters of \textsc{Imsbs} are tuned on the set \textsc{Random} using \texttt{irace}~\cite{perez2014analysis}, with a total budget of 5000 runs (each limited to 30 minutes). For tuning, one instance from each group is selected, involving 32 training instances in total. All remaining (less sensitive) parameters are fixed to the values reported  in~\cite{djukanovic2026_eurocast}. The produced configuration for \textsc{Imsbs} is $\beta=2000$, $h=\text{UB}$, and $\#\texttt{imbs\_iters}=5000$.
 
For a fair comparison, the proposed \textsc{Limsbs-ensemble} inherits the same parameter configuration obtained for \textsc{Imsbs}. The learning process is tuned by exploring different neural network architectures: the multi-layer perceptrons which combine two or three layers by  allowing 5 or 10 neurons are considered. One learning process per each such architecture is executed. To reduce the computational burden, we restrict the activation function to be identical across all layers, selecting from the following three: $\{\texttt{tanh}, \texttt{ReLU}, \texttt{sigmoid}\}$. To analyze the sensitivity with respect to feature design, we define three feature configurations: $i$) $F_1$: 4 features from sequence positions, 4 features from gap constraints, and the subsequence length; $ii$) $F_2$: $F_1$ extended with the alphabet size $|\Sigma|$; $iii$) $F_3$: $F_2$ extended with the number of sequences $m$.

After extensive empirical evaluation, the following NN architectures are selected for the \textsc{Random} benchmark:
\begin{itemize}
	\item $m=2$: a 3-layer neural network with  10, 5, and 5 neurons at respective layers, apart from one corresponding to the bias;  each layer incorporates the activation \texttt{ReLU}, with the feature set  $F_2$;
	\item $m=3$:  a 3-layer neural network with  10, 10, and 5 neurons at respective layers, apart from one corresponding to the bias;    each layer incorporates the activation \texttt{ReLU}, with the feature set  $F_2$;
	\item $m=5$:   a 3-layer neural network with  10, 5, and 5 neurons at respective layers, apart from one corresponding to the bias;   each layer incorporates the activation \texttt{sigmoid}, with the feature set  $F_1$;  
	\item $m=10$: a 3-layer neural network with  10, 5, and 5 neurons at respective layers, apart from one corresponding to the bias;    each layer incorporates the activation \texttt{ReLU}, with the feature set  $F_2$.
\end{itemize}

\begin{figure}[!ht]
	\centering
	\begin{minipage}{0.48\textwidth}
		\centering
		\includegraphics[width=\linewidth,height=90pt]{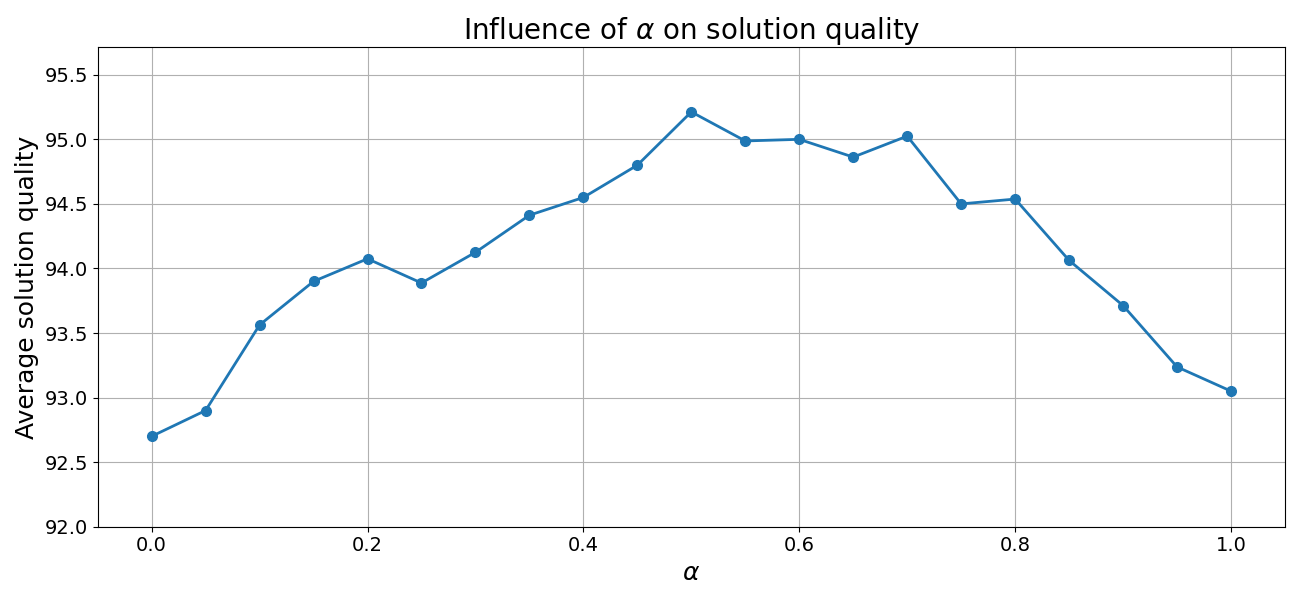}
	\end{minipage}
	\hfill
	\begin{minipage}{0.48\textwidth}
		\centering
		\includegraphics[width=\linewidth,height=90pt]{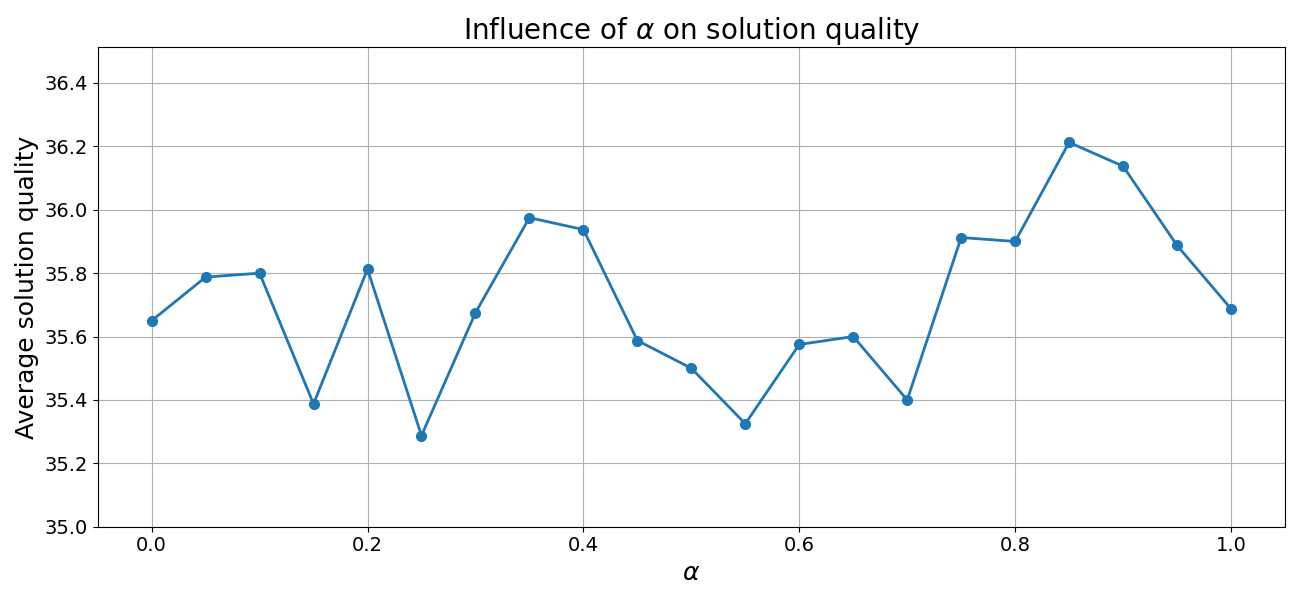}
	\end{minipage}
	
	\caption{Visualization of results for tuning parameter $\alpha$: $m=3$  (left), and $m=5$ (right).}
	\label{fig:alpha_tuning}
\end{figure}

For training the neural heuristic, we use 32 instances for the training $\mathcal{T}$ and 32 instances for the validation set $\mathcal{V}$, selecting one instance per each group. For each value of $m$, this results in 8 instances used in the training phase and 8 instances in the validation phase.   After the heuristic $h_w$ is learned, the ensemble parameter is set to $\alpha = 0.6$ for $m=2$, $\alpha = 0.5$ for $m=3$, $\alpha = 0.35$ for $m=5$, and $\alpha = 0.15$ for $m=10$, based on grid search, see Figure~\ref{fig:alpha_tuning}.


For the \textsc{Real} benchmark, we employ the same NN architecture as in the $m=10$ case for the \textsc{Random} benchmark set. This choice allows us to asses the robustness of the learned heuristic on structurally different instances with a larger number of sequences ($m \geq 10$). The \textsc{Imsbs} parameters remain unchanged, and follows those used to test on the set \textsc{Random}, while we set $\alpha=0.65$ according to grid search performed on benchmark set \textsc{Real}. 

\subsection{Results on Benchmark Set \textsc{Random}}

\begin{table}[t]
	\centering
	\caption{Comparison of objective values between \textsc{Imsbs} and \textsc{Limsbs}-ensemble on benchmark set \textsc{Random}. Best results per instance category are displayed in bold.}
	\label{tab:preliminary_results}
	
	\begin{minipage}{0.4\textwidth}
		\centering
		\setlength{\tabcolsep}{4pt}
		\renewcommand{\arraystretch}{1.1}
		\scalebox{0.7}{
			\begin{tabular}{cll|ll}
				\hline
				$m$ & $n$ & $|\Sigma|$ & \textsc{Limsbs}-ensemble & \textsc{Imsbs} \\ \hline
2 &  50 & 2 & \textbf{37.7} & \textbf{37.7} \\
2 &  50 & 4 & \textbf{30.1} & \textbf{30.1} \\
 2 & 100 & 2 & 72.4 & \textbf{72.9} \\
2 & 100 & 4 & \textbf{62.1} & 61.9 \\
2 & 200 & 2 & \textbf{140.7} & 139.3 \\
2 & 200 & 4 & \textbf{125.8} & 125.5 \\
2 & 500 & 2 & \textbf{299.5} & 282.1 \\
2 & 500 & 4 & \textbf{310.9} & 303.0 \\ \hline
3 &  50 & 2 & \textbf{31.2} & \textbf{31.2} \\
3 &  50 & 4 & \textbf{22.9} & \textbf{22.9} \\
3 & 100 & 2 & \textbf{58.5} & \textbf{58.5} \\
3 & 100 & 4 & \textbf{48.7} & 48.5 \\
3 & 200 & 2 & \textbf{106.9} & 106.4 \\
3 & 200 & 4 & \textbf{98.0} & 97.8 \\
3 & 500 & 2 & \textbf{157.2} & 150.6 \\
3 & 500 & 4 & \textbf{238.3} & 236.0 \\
				\hline \hline
Avg. &  &  & \textbf{115.1}  & {112.8} \\ \hline \hline
		\end{tabular}}
	\end{minipage}
	\hfill
	\begin{minipage}{0.4\textwidth}
		\centering
		\setlength{\tabcolsep}{4pt}
		\renewcommand{\arraystretch}{1.1}
		\scalebox{0.7}{
			\begin{tabular}{cll|ll}
				\hline
				$m$ & $n$ & $|\Sigma|$ & \textsc{Limsbs}-ensemble & \textsc{Imsbs} \\
				\hline
 5 &  50 & 2 & \textbf{20.5} & \textbf{20.5} \\
5 &  50 & 4 & \textbf{15.3} & \textbf{15.3} \\
 5 & 100 & 2 & 32.5 & \textbf{32.9} \\
 5 & 100 & 4 & 27.6 & \textbf{28.1} \\
 5 & 200 & 2 & \textbf{47.3} & 42.1 \\
 5 & 200 & 4 & \textbf{41.2} & 38.8 \\
 5 & 500 & 2 & \textbf{46.0} & 39.2 \\
 5 & 500 & 4 & \textbf{59.3} & 45.6 \\ \hline
 10 &  50 & 2 & {11.6} & \textbf{11.7} \\
 10 &  50 & 4 & \textbf{7.5} & \textbf{7.5} \\
 10 & 100 & 2 & \textbf{14.4} & 12.0 \\
 10 & 100 & 4 & \textbf{9.9} & 8.9 \\
 10 & 200 & 2 & \textbf{14.8} & 12.8 \\
 10 & 200 & 4 & \textbf{9.5} & 8.2 \\
 10 & 500 & 2 & \textbf{14.6} & 13.2 \\
 10 & 500 & 4 & \textbf{10.1} & 8.6 \\
				\hline \hline
Avg. &    &   &   \textbf{23.8}    & 21.6 \\ \hline \hline
		\end{tabular}}
	\end{minipage}

\end{table}

Table~\ref{tab:preliminary_results} reports the average results over 10 instances per group for both competitor algorithms. The table is organized into two parts: instance characteristics given by the first three columns followed by two columns reporting the performance metrics of \textsc{Limsbs-ensemble} and \textsc{Imsbs}, respectively.

To assess statistical significance, we perform a one-sided  Wilcoxon signed-rank test on paired instances (see Table~\ref{tab:statistical_test}).

\begin{table}[!ht]
	\centering
	\caption{Statistical comparison between \textsc{Limsbs-ensemble} and \textsc{Imsbs}.}
	\label{tab:statistical_test}
	\scalebox{0.8}{
		\begin{tabular}{lccc}
			\hline
			\textbf{Comparison} & \textbf{Wins / Ties / Losses} & \textbf{p-value} & \textbf{Conclusion} \\
			\hline
			All instances ($m = 2,3,5,10$) & 20 / 8 / 4 & 0.009 & Significant improvement \\
			Small instances ($m = 2,3$)    & 10 / 5 / 1  & $<0.01$ & Significant improvement \\
			Large instances ($m = 5,10$)   & 10 / 3 / 3 & $< 0.05$ & Significant improvement \\
			\hline
	\end{tabular}}

\end{table}

\begin{figure}[!ht]
	\centering
	\includegraphics[width=300pt,height=140pt]{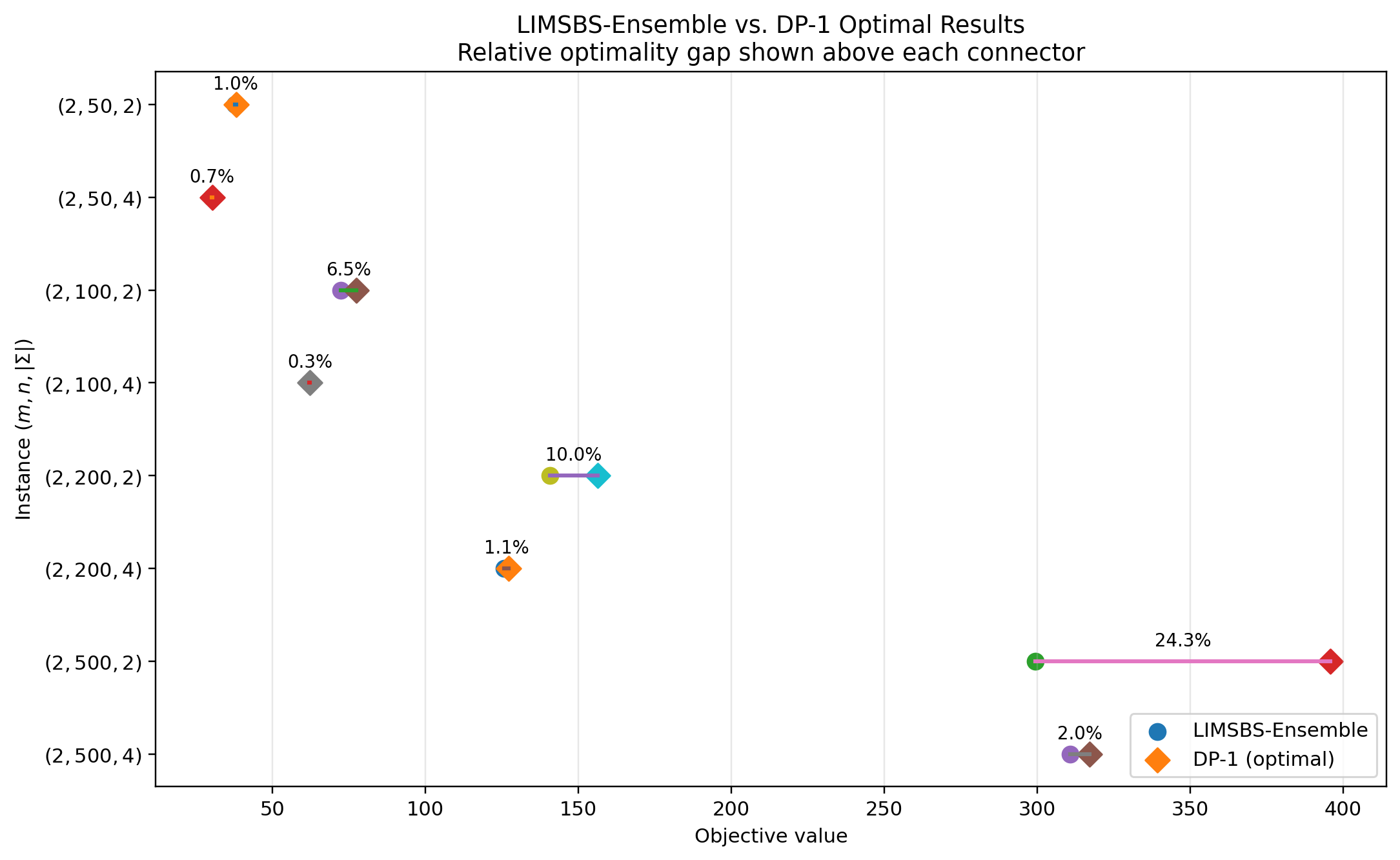}
	\caption{Comparison to known optimal results: $m=2$ case.}\label{fig:limsbs_vs_dp1}
\end{figure}

The following conclusions are drawn from these results: $i)$  Out of 32 instances, \textsc{Limsbs-ensemble} achieves better performance over \textsc{Imsbs} in 20 cases, ties occur in 8 cases, and in 4 cases the winner is \textsc{Imsbs}; $ii$) The performed statistical evaluation suggests that there is a statistically significant improvement of \textsc{Limsbs-ensemble} over \textsc{Imsbs}.
Additionally, in Figure~\ref{fig:limsbs_vs_dp1} we show the relative gaps between the results of \textsc{Limsbs-Ensemble} and the exact solutions of dynamic programming (DP-1) approach from literature~\cite{djukanovic2026_eurocast}. While the results of \textsc{Limsbs-Ensemble} are tight for $|\Sigma|=4$, it seems there is still room for improvements  in the case of instances with $|\Sigma|=2$.

\subsection{Results on Benchmark Set \textsc{Real}}

\begin{table}[!ht]
	\centering
	\caption{Comparison of \textsc{Limsbs}-ensemble and \textsc{Imsbs} on benchmark set \textsc{Real}.}
	\label{tab:real_results}
	
	\begin{minipage}{0.45\textwidth}
		\centering
		\textbf{RAT instances}
		\scalebox{0.7}{
		\begin{tabular}{l|lr|lr|r}
			\hline
			$m$ & \textsc{Limsbs}-ensemble & ${t}[s]$ & \textsc{Imsbs} & ${t}[s]$ & LCS$^*$ \\
			\hline \hline
			10 & \textbf{26} & 69.58 & {25}          & 30.65 & 209 \\
			15 & \textbf{12} & 40.81 & {11} & 24.26 & 189    \\  
			20 & \textbf{13} & 16.90 & 9           & 49.36 & 174 \\
			
			25 & \textbf{7}  & 24.30 & \textbf{7}          & 29.62 & 173   \\
			40 & \textbf{5}  & 19.25 & 4           & 48.64 & 154  \\
			
			60 & \textbf{3}  & 25.81 & 1           & 45.22 & 154   \\
			
			80 & \textbf{1}  & 27.10 & \textbf{1}  & 53.02 & 144   \\
			
			100 & \textbf{1}  & 55.98 & \textbf{1}  & 44.30 & 139  \\

			150 & \textbf{1}  & 56.73 & \textbf{1}  & 66.57 & 131  \\

			200 & \textbf{1}  & 38.84 & \textbf{1}  & 62.54 & 126 \\
  
			\hline
		\end{tabular}}
	\end{minipage}
	\hfill \hfill
	\begin{minipage}{0.45\textwidth}
		\centering
		\textbf{VIRUS instances}
		
		\scalebox{0.7}{
		  \begin{tabular}{l|lr|lr|r}
			\hline
			$m$ & \textsc{Limsbs}-ensemble & ${t}[s]$ & \textsc{Imsbs} & ${t}[s]$& LCS$^*$ \\
			\hline \hline		
			10 & \textbf{84} & 163.59 & 36         & 43.14 &  228 \\
		
			15 & {18} & 25.21 & \textbf{19} & 36.52 &  206 \\
			
			20 & \textbf{12} & 24.13 & {10}          & 26.92 &  194 \\
			
			25 & \textbf{12} & 25.17 & 9           & 30.40 &  196 \\
			
			40 & \textbf{7}  & 23.41 & 5           & 32.12 &  174 \\
			
			60 & \textbf{5}  & 25.95 & 4           & 34.38 &  168 \\
			
			80 & \textbf{4}  & 34.47 & 3           & 40.64 &  163 \\
			
			100 & \textbf{3}  & 55.68 & \textbf{3}  & 43.25 & 160 \\
			
			150 & \textbf{4}  & 42.87 &  {3}           & 31.80 & 157 \\
			
			200 & \textbf{1}  & 38.64 & \textbf{1}  & 48.59 & 156  \\	\hline
		\end{tabular}}
	\end{minipage}
	\vspace{-0.5cm}
\end{table}
Table~\ref{tab:real_results} summarizes the results, including solution quality and runtime ($t$ in seconds) for both approaches; additionally, the reference (best-known) LCS values from the literature (LCS$^*$) are also reported to reveal the effect of gap constraints on constraint-free problem version. The following observations are drawn.

\begin{itemize}
	\item \textit{Solution quality.} \textsc{Limsbs-ensemble} outperforms \textsc{Imsbs} in 12 out of 20 instances, with 7 cases resulting in ties, and only one in favor of \textsc{Imsbs}. 
	
	
	\item \textit{Instance difficulty.} The improvements of \textsc{Limsbs-ensemble} over \textsc{Imsbs} are more pronounced on challenging instances (e.g., \textsc{Virus} dataset), where substantial gains are observed in several occasions. Ties typically occur on more constrained instances (those with a large number of constraints $m$) with low objective values.
	
	\item \textit{Runtime.} The computational effort of both approaches is comparable. While \textsc{Imsbs} is slightly faster on average, the differences are not consistent, and \textsc{Limsbs-ensemble} is competitive in most cases.
	
	\item \textit{Impact of gap constraints.} Compared to unconstrained LCS values from the literature (reported in the last columns of both tables), introducing gap constraints significantly reduces achievable subsequence lengths. This effect is particularly strong for large $m$ (e.g., $m \geq 60$), where feasible solutions become highly restricted, the search space reduced, while the number of root nodes exponentially  increases with the size of $m$. 
\end{itemize}

Overall, the results demonstrate that the novel \textsc{Limsbs-ensemble}   provides a consistent improvement in solution quality over \textsc{Imsbs}, while maintaining comparable computational efficiency. This makes it a preferable choice when solution quality is the primary objective in addressing the tackled problem.

\section{Conclusions}\label{sec:conclusions}

This paper addressed the Variable Gapped Longest Common Subsequence Problem (VGLCSP), a challenging extension of the classical LCSP with applications in biological sequence analysis and time-series modeling. We proposed a data-driven learning-based heuristic designed to guide the search process within the state-of-the-art iterative multi-source beam search (\textsc{Imsbs}) framework.

The heuristic is represented by a neural network with a predefined architecture, whose parameters are optimized via a biased random-key genetic algorithm in a neuro-evolutionary setting. To further enhance robustness, we introduced an ensemble heuristic that combines the learned heuristic with the best-performing hand-crafted heuristic from the literature. This hybrid guidance mechanism is integrated into \textsc{Imsbs} to improve search efficiency and solution quality. Experimental results on both synthetic benchmarks and newly introduced biologically inspired instances demonstrate that the proposed ensemble heuristic  significantly outperforms the baseline hand-crafted heuristics integrated into \textsc{Imsbs}. In particular, the improvements are most pronounced on challenging instances with complex gap structures, confirming the effectiveness of the learning-guided search.  

Future work will focus on extending the approach to protein sequence datasets and larger-scale instances to further assess the scalability. Another aspect is validation of gap constraints from our real-world generated instances and their biological context. One could compare with known sequence-alignment benchmarks.  Additionally, alternative learning strategies could be explored, including different evolutionary schemes, fitness evaluation criteria beyond mean performance, and improved stopping mechanisms to better control overfitting.\\
 \footnotesize{\textbf{Acknowledgments.}    This publication is co- funded by the European Union’s Horizon Europe research and innovation program under the Marie Sklodowska-Curie COFUND Postdoctoral Programme grant agreement No.101081355-- SMASH and by the Republic of Slovenia and the European Union from the European Regional Development Fund. Co-funded by European Union.  Views and opinions expressed are those of the authors only and do not necessarily reflect those of the European Union or European Research Executive Agency (REA). Neither the European Union nor the REA can be held responsible for them.  The authors gratefully acknowledge the SLING consortium for funding this research by providing computing resources of the HPC Vega at the Institute of Information Science (www.izum.si).}

	\bibliographystyle{splncs04}
	\bibliography{bib}
	%
	
	
	
\end{document}